# Efficient Deep Processing of Japanese

Melanie SIEGEL
DFKI GmbH
Stuhlsatzenhausweg 3
66123 Saarbrücken, Germany
siegel@dfki.de

Emily M. BENDER
CSLI Stanford
220 Panama Street
Stanford, CA, 94305-4115, USA
bender@csli.stanford.edu

**Abstract**

We present a broad coverage Japanese grammar written in the HPSG formalism with MRS semantics. The grammar is created for use in real world applications, such that robustness and performance issues play an important role. It is connected to a POS tagging and word segmentation tool. This grammar is being developed in a multilingual context, requiring MRS structures that are easily comparable across languages.

**Introduction**

Natural language processing technology has recently reached a point where applications that rely on deep linguistic processing are becoming feasible. Such applications (e.g. message extraction systems, machine translation and dialogue understanding systems) require natural language understanding, or at least an approximation thereof. This, in turn, requires rich and highly precise information as the output of a parse. However, if the technology is to meet the demands of real-world applications, this must not come at the cost of robustness. Robustness requires not only wide coverage by the grammar (in both syntax and semantics), but also large and extensible lexica as well as interfaces to preprocessing systems for named entity recognition, non-linguistic structures such as addresses, etc. Furthermore, applications built on deep NLP technology should be extensible to multiple languages. This requires flexible yet well-defined output structures that can be adapted to grammars of many different languages. Finally, for use in real-world applications, NLP systems meeting the above desiderata must also be efficient.

In this paper, we describe the development of a broad coverage grammar for Japanese that is used in an automatic email response application. The grammar is based on work done in the *Verbmobil* project (Siegel 2000) on machine translation of spoken dialogues in the domain of travel planning. It has since been greatly extended to accommodate written Japanese and new domains.

The grammar is couched in the theoretical framework of Head-Driven Phrase Structure Grammar (HPSG) (Pollard & Sag 1994), with semantic representations in Minimal Recursion Semantics (MRS) (Copestake et al. 2001). HPSG is well suited to the task of multilingual development of broad coverage grammars: It is flexible enough (analyses can be shared across languages but also tailored as necessary), and has a rich theoretical literature from which to draw analyzes and inspiration. The characteristic type hierarchy of HPSG also facilitates the development of grammars that are easy to extend. MRS is a flat semantic formalism that works well with typed feature structures and is flexible in that it provides structures that are under-specified for scopal information. These structures give compact representations of ambiguities that are often irrelevant to the task at hand.

HPSG and MRS have the further advantage that there are practical and useful open-source tools for writing, testing, and efficiently processing grammars written in these formalisms. The tools we are using in this project include the LKB system (Copestake 2002) for grammar development, [incr tsdb()] (Oepen & Carroll 2000) for testing the grammar and tracking changes, and PET (Callmeier 2000), a very efficient HPSG parser, for processing. We also use the ChaSen tokenizer and POS tagger (Asahara & Matsumoto 2000).

While couched within the same general framework (HPSG), our approach differs from that of Kanayama et al (2000). The work described there achieves impressive coverage (83.7% on the EDR corpus of newspaper text) with an underspecified grammar consisting of a small number of lexical entries, lexical types associated with parts of speech, and six underspecified grammar rules. In contrast, our grammar is much larger in terms of the number of lexical entries, the number of grammar rules, and the constraints on both,[1] and takes correspondingly more effort to bring up to that level of coverage. The higher level of detail allows us to output precise semantic representations as well as to use syntactic, semantic and lexical information to reduce ambiguity and rank parses.

# 1   Japanese HPSG Syntax

The fundamental notion of an HPSG is the sign. A sign is a complex feature structure representing information of different linguistic levels of a phrase or lexical item. The attribute-value matrix of a sign in the Japanese HPSG is quite similar to a sign in the LinGO English Resource Grammar (henceforth ERG) (Flickinger 2000), with information about the orthographical realization of the lexical sign in PHON, syntactic and semantic information in SYNSEM, information about the lexical status in LEX, nonlocal information in NONLOC, head information that goes up the tree in HEAD and information about subcategorization in SUBCAT.

The grammar implementation is based on a system of types. There are 900 lexical types that define the syntactic, semantic and pragmatic properties of the Japanese words, and 188 types that define the properties of phrases and lexical rules. The grammar includes 50 lexical rules for inflectional and derivational morphology and 47 phrase structure rules. The lexicon contains 5100 stem entries. As the grammar is developed for use in applications, it treats a wide range of basic constructions of Japanese. Only some of these phenomena can be described here.

## 1.1   Subcategorization

The structure of SUBCAT is different from the ERG SUBCAT structure. This is due to differences in subcategorization between Japanese and English. A fundamental difference is the fact that, in Japanese, verbal arguments are frequently omitted. For example, arguments that refer to the speaker, addressee, and other arguments that can be inferred from context are often omitted in spoken language. Additionally, optional verbal arguments can scramble. On the other hand, some arguments are not only obligatory, but must also be realized adjacent to the selecting head.

To account for this, our subcategorization contains the attributes SAT and VAL. The SAT value encodes whether a verbal argument is already saturated (such that it cannot be saturated again), optional or adjacent. VAL contains the agreement information for the argument. When an argument is realized, its SAT value on the mother node is specified as *sat* and its SYNSEM is unified with its VAL value on the subcategorizing head. The VAL value on the mother is *none*. Adjacency must be checked in every rule that combines heads and arguments or adjuncts. This is the *principle of adjacency*, stated as follows:

*In a headed phrase, the SUBCAT.SAT value on the non-head daughter must not contain any adjacent arguments. In a head-complement structure, the SUBCAT.SAT value of the head daughter must not contain any adjacent arguments besides the non-head daughter. In a head-adjunct structure, the SUBCAT.SAT value of the head daughter must not contain any adjacent arguments.*

## 1.2   Verbal inflection

Japanese verb stems combine with endings that provide information about honorification, tense, aspect, voice and mode. Inflectional rules for the different types of stems prepare the verb stems for combination with the verbal endings. For example, the verb stem *yomu* must be inflected to *yon* to combine with the past tense ending *da*. Morphological features constrain the combination of stem and ending. In the above example, the inflectional rule changes the *mu* character to the *n* character and assigns the value *nd-morph* to the morphological feature `RMORPH-BIND-TYPE`. The ending *da* selects for a verbal stem with this value.

---

[1] We do also make use of generic lexical entries for certain parts of speech as a means of extending our lexicon. See section 3 below.

Endings can be combined with other endings, as in *-sase-rare-mashi-ta* (causative-potential-honorific-past), but not arbitrarily:

    **-sase-mashi-rare-ta*
    **-sase-ta-mashi-rare*
    *-sase-ta*
    *-rare-mashi-ta*

This is accounted for with two kinds of rules which realize mutually selected elements. In the combination of stem and ending, the verb stem selects for the verbal ending via the head feature `SPEC`. In the case of the combination of two verbal endings, the first ending selects for the second one via the head feature `MARK`. In both cases, the right element subcategorizes for the left one via `SUBCAT.VAL.SPR`. Using this mechanism, it is possible to control the sequence of verbal endings: Verb stems select verbal endings via `SPEC` and take no `SPR`, derivational morphemes (like causative or potential) select tense endings or other derivational morphemes via `MARK` and subcategorize for verb stems and/or verb endings via `SPR` (*sase* takes only verb stems), and tense endings take verb stems or endings as `SPR` and take no `MARK` or `SPEC` (as they occur at the end of the sequence).

## 1.3 Complex Predicates

A special treatment is needed for Japanese verbal noun + light verb constructions. In these cases, a word that combines the qualities of a noun with those of a verb occurs in a construction with a verb that has only marginal semantic information. The syntactic, semantic and pragmatic information on the complex is a combination of the information of the two.

Consider example 1. The verbal noun *benkyou* contains subcategorization information (transitive), as well as semantic information (the *benkyou*-relation and its semantic arguments). The light verb *shi-ta* supplies tense information (past). Pragmatic information can be supplied by both parts of the construction, as in the formal form *o-benkyou shi-mashi-ta*. The rule that licenses this type of combination is the *vn-light-rule*, a subtype of the *head-marker-rule*.

Example 1:
*Benkyou shi-ta.*
```
study   do-past
```
*'Someone has studied.'*

Japanese auxiliaries combine with verbs and provide either aspectual or perspective information or information about honorification. In a verb-auxiliary construction, the information about subcategorization is a combination of the `SUBCAT` information of verb and auxiliary, depending on the type of auxiliary. The rule responsible for the information combination in these cases is the *head-specifier-rule*. We have three basic types of auxiliaries. The first type is aspect auxiliaries. These are treated as raising verbs, and include such elements as *iru* (roughly, progressive) and *aru* (roughly, perfective), as can be seen in example 2. The other two classes of auxiliaries provide information about perspective or the point of view from which a situation is being described. Both classes of auxiliaries add a *ni* (dative) marked argument to the argument structure of the whole predicate. The classes differ in how they relate their arguments to the arguments of the verb. One class (including *kureru* 'give'; see example 3) are treated as subject control verbs. The other class (including *morau* 'receive', see example 4) establishes a control relation between the *ni*-marked argument and the embedded subject.

Example 2:
*Keeki wo tabe-te iru.*
```
cake ACC eat  progressive
```
*'Someone is eating cake.'*

Example 3:
*Sensei  wa   watashi ni   hon   wo*
```
teacher TOP  I       DAT  book  ACC
```
*katte kure-ta.*
```
buy   give-past
```
*'The teacher bought me a book.'*

Example 4:
*Watashi ga  sensei  ni   hon    wo*
```
I       NOM teacher DAT  book   ACC
```
*katte   morat-ta.*
```
buy     get-past
```
*'The teacher bought me a book.'*

### 1.4 Particles in a type hierarchy

The careful treatment of Japanese particles is essential, because they are the most frequently occurring words and have various central functions in the grammar. It is difficult, because one particle can fulfill more than one function and they can co-occur, but not arbitrarily. The Japanese grammar thus contains a type hierarchy of 44 types for particles. See Siegel (1999) for a more detailed description of relevant phenomena and solutions.

### 1.5 Numeral Expressions

Number names, such as *sen kyuu hyaku juu* '1910' constitute a notable exception to the general head-final pattern of Japanese phrases. We found Smith's (1999) head-medial analysis of English number names to be directly applicable to the Japanese system as well (Bender 2002). This analysis was easily incorporated into the grammar, despite the oddity of head positioning, because the type hierarchy of HPSG is well suited to express the partial generalizations that permeate natural language.

On the other hand, number names in Japanese contrast sharply with number names in English in that they are rarely used without a numeral classifier.

Example 5:
*Juu  *(hiki  no)  neko ga  ki-ta.*
ten   CL   GEN  cat  NOM arrive-past
*'Ten cats arrived.'*

The grammar provides for 'true' numeral classifiers like *hon*, *ko*, and *hiki*, as well as formatives like *en* 'yen' and *do* 'degree' which combine with number names just like numeral classifiers do, but never serve as numeral classifiers for other nouns. In addition, there are a few non-branching rules that allow bare number names to surface as numeral classifier phrases with specific semantic constraints.

### 1.6 Pragmatic information

Spoken language and email correspondence both encode references to the social relation of the dialogue partners. Utterances can express social distance between addressee and speaker and third persons. Honorifics can even express respect towards inanimates. Pragmatic information is treated in the CONTEXT layer of the complex signs. Honorific information is given in the CONTEXT.BACKGROUND and linked to addressee and speaker anchors.

The expression of empathy or in-group vs. out-group is quite prevalent in Japanese. One means of expressing empathy is the perspective auxiliaries discussed above. For example, two auxiliaries meaning roughly 'give' (*ageru* and *kureru*) contrast in where they place the empathy. In the case of *ageru*, it is with the giver. In the case of *kureru*, it is with the recipient. We model this within the sign by positing a feature EMPATHY within CONTEXT and linking it to the relevant arguments' indices.

## 2 Japanese MRS Semantics

In the multilingual context in which this grammar has been developed, a high premium is placed on parallel and consistent semantic representations between grammars for different languages. Ensuring this parallelism enables the reuse of the same downstream technology, no matter which language is used as input. Integrating MRS representations parallel to those used in the ERG into the Japanese grammar took approximately 3 months. Of course, semantic work is on-going, as every new construction treated needs to be given a suitable semantic representation. For the most part, semantic representations developed for English were straightforwardly applicable to Japanese. This section provides a brief overview of those cases where the Japanese constructions we encountered led to innovations in the semantic representations and/or the correspondence between syntactic and semantic structures. Due to space limitations, we discuss these analyses in general terms and omit technical details.

### 2.l Nominalization and Verbal Nouns

Nominalization is of course attested in English and across languages. However, it is much more prevalent in Japanese than in English, primarily because of verbal nouns. As noted in Section 1.3 above, a verbal noun like *benkyou* 'study' can appear in syntactic contexts requiring nouns, or, in combination with a light verb, in contexts requiring verbs. One possible analysis would provide two separate lexical entries, one with nominal and one with verbal semantics. However, this would not only be redundant (missing the systematic relationship between these uses of verbal nouns) but would also contradict the intuition that even in its nominal use, the arguments of *benkyou* are still present.

Example 6:
*Nihongo    no    benkyou wo  hajimeru.*
```
Japanese GEN study   ACC begin
```
*'Someone begins the study of Japanese.'*

In order to capture this intuition, we opted for an analysis that essentially treats verbal nouns as underlyingly verbal.  The nominal uses are produced by a lexical rule which nominalizes the verbal nouns.  The semantic effect of this rule is to provide a nominal relation which introduces a variable which can in turn be bound by quantifiers. The nominal relation subordinates the original verbal relation supplied by the verbal noun.  The rule is lexical as we have not yet found any cases where the verb's arguments are clearly filled by phrases in the syntax.  If they do appear, it is with genitive marking (e.g., *nihongo no* in the example above).  In order to reduce ambiguity, we leave the relationship between these genitive marked NPs and the nominalized verbal noun underspecified.  There is nothing in the syntax to disambiguate these cases, and we find that they are better left to downstream processing, where there may be access to world knowledge.

## 2.2   Numeral Classifiers

As noted in Section 1.5, the internal syntax of number names is surprisingly parallel between English and Japanese, but their external syntax differs dramatically. English number names can appear directly as modifiers of NPs and are treated semantically as adjectives in the ERG. Japanese number names can only modify nouns in combination with numeral classifiers. In addition, numeral classifier phrases can appear in NP positions (akin to partitives in English). Finally, some numeral-classifier-like elements do not serve the modifier function but can only head phrases that fill NP positions.

   This constellation of facts required the following innovations: a representation of numbers that doesn't treat them as adjectives (in MRS terms, a feature structure without the ARG feature), a representation of the semantic contribution of numeral classifiers (a relation between numbers and the nouns they modify, this time with an ARG feature), and a set of rules for promoting numeral classifier phrases to NPs that contribute the appropriate nominal semantics (underspecified in the case of ordinary numeral classifiers or specific in the case of words like *en* 'yen').

## 2.3   Relative Clauses and Adjectives

The primary issue in the analysis of relative clauses and adjectives is the possibility of extreme ambiguity, due to several intersecting factors:  Japanese has rampant pro-drop and does not have any relative pronouns.  In addition, a head noun modified by a relative clause need not correspond to any gap in the relative clause, as shown by examples like the following (Matsumoto 1997):

Example 7:
*atama    ga    yoku    naru    hon*
```
head    NOM better become book
```
*'a book that makes one smarter'*

Therefore, if we were to posit an attributive adjective + noun construction (distinct from the relative clause + noun possibility) we would have systematic ambiguities for NPs like *akai hon* ('red book'), ambiguities which could never be resolved based on information in the sentence.  Instead, we have opted for a relative clause analysis of any adjective + noun combination in which the adjective could potentially be used predicatively.  Furthermore, because of gapless relative clauses like the one cited above, we have opted for a non-extraction analysis of relative clauses.[2]

   Nonetheless, the well-formedness constraints on MRS representations require that there be some relationship between the head noun and the relative clause. We picked the topic relation for this purpose (following Kuno 1973). The topic relation is introduced into the semantics by the relative clause rule. As with main clause topics (which we also give a non-extraction analysis), we rely on downstream anaphora resolution to refine the relationship.

---

[2] There is in fact some linguistic evidence for extraction in some relative clauses in Japanese  (see e.g., Baldwin 2001).  However, we saw no practical need to allow for this possibility in our grammar, and particularly not one that would justify the increase in ambiguity.  There is also evidence that some adjectives are true attributives and cannot be used predicatively (Yamakido 2000). These are handled by a separate adjective + noun rule restricted to just these cases.

## 2.4 Summary

For the most part, semantic representations and the syntax-semantic interface already worked out in the ERG were directly applicable to the Japanese grammar. In those cases where Japanese presented problems not yet encountered (or at least not yet tackled) in English, it was fairly straightforward to work out suitable MRS representations and means of building them up. Both of these points illustrate the cross-linguistic validity and practical utility of MRS representations.

## 3 Integration of a Morphological Analyzer

As Japanese written text does not have word segmentation, a preprocessing system is required. We integrated ChaSen (Asahara & Matsumoto 2000), a tool that provides word segmentation as well as POS tags and morphological information such as verbal inflection. As the lexical coverage of ChaSen is higher than that of the HPSG lexicon, default part-of-speech entries are inserted into the lexicon. These are triggered by the part-of-speech information given by ChaSen, if there is no existing entry in the lexicon. These specific default entries assign a type to the word that contains features typical to its part-of-speech. It is therefore possible to restrict the lexicon to those cases where the lexical information contains more than the typical information for a certain part-of-speech. This default mechanism is often used for different kinds of names and 'ordinary' nouns, but also for adverbs, interjections and verbal nouns (where we assume a default transitive valence pattern).[3]

---

[3] Kanayama et al. (2000) use a similar mechanism for most words. They report only 105 grammar-inherent lexical entries.

The ChaSen lexicon is extended with a domain-specific lexicon, containing, among others, names in the domain of banking.

For verbs and adjectives, ChaSen gives information about stems and inflection that is used in a similar way. The inflection type is translated to an HPSG type. These types interact with the inflectional rules in the grammar such that the default entries are inflected just as 'known' words would be.

In addition to the preprocessing done by ChaSen, an additional (shallow) preprocessing tool recognizes numbers, date expressions, addresses, email addresses, URLs, telephone numbers and currency expressions. The output of the preprocessing tool replaces these expressions in the string with placeholders. The placeholders are parsed by the grammar using special placeholder lexical entries.

## 4 Robustness and Performance Issues

The grammar is aimed at working with real-world data, rather than at experimenting with linguistic examples. Therefore, robustness and performance issues play an important role. While grammar development is carried out in the LKB (Copestake 2002), processing (both in the application domain and for the purposes of running test suites) is done with the highly efficient PET parser (Callmeier 2000). Figures 1 and 2 show the performance of PET parsing of hand-made and real data, respectively.

| *Phenomenon* | *items* # | *etasks* Ø | *filter* % | *edges* Ø | *first* Ø (s) | *total* Ø (s) | *tcpu* Ø (s) | *gc* Ø (s) | *space* Ø (kb) |
|---|---|---|---|---|---|---|---|---|---|
| *Total* | *742* | *946* | *95.7* | *303* | *0.06* | *0.11* | *0.11* | *0* | *833* |

Fig.1 Performance parsing banking data, generated by [incr tsdb()]

| *Phenomenon* | *items* # | *etasks* Ø | *filter* % | *edges* Ø | *first* Ø (s) | *total* Ø (s) | *tcpu* Ø (s) | *tgc* Ø (s) | *space* Ø (kb) |
|---|---|---|---|---|---|---|---|---|---|
| *Total* | *316* | *2020* | *96.5* | *616* | *0.23* | *0.26* | *0.26* | *0* | *1819* |

Fig.2 Performance parsing document request data, generated by [incr tsdb()]

One characteristic of real-world data is the variety of punctuation marks that occur and the potential for ambiguity that they bring. In our grammar, certain punctuation marks are given lexical entries and processed by grammar rules. Take, for example, quotation marks. Ignoring them (as done in most development-oriented grammars and smaller grammars), leads to a significant loss of structural information:

Example 8:
*"Botan    wo    osu"    to          it-ta*
`button ACC push COMPL say-past`
'Someone said: "push the button."'

The formative *to* is actually ambiguous between a complementizer and a conjunction. Since the phrase before *to* is a complete sentence, this string is ambiguous if one ignores the quotation marks. With the quotation marks, however, only the complementizer *to* is possible. Given the high degree of ambiguity inherent in broad-coverage grammars, we have found it extremely useful to parse punctuation rather than ignore it.

The domains we have been working on (like many others) contain many date and number expressions. While a shallow tool recognizes general structures, the grammar contains rules and types to process these.

Phenomena occurring in semi-spontaneous language (email correspondence), such as interjections (e.g. *maa* 'well'), contracted verb forms (e.g. *tabe-chatta* < *tabete-shimatta* '(someone) ate it all up'), fragmentary sentences (e.g. *bangou: 1265* 'number: 1265') and NP fragments (e.g. *bangou*? 'number?') must be covered as well as the 'ordinary' complete sentences found in more carefully edited text. Our grammar includes types, lexical entries, and grammar rules for dealing with such phenomena.

Perhaps the most important performance issue for broad coverage grammars is ambiguity. At one point in the development of this grammar, the average number of readings doubled in two months of work. We currently have two strategies for addressing this problem: First, we include a mechanism into the grammar rules that chooses left-branching rules in cases of compounds, genitive modification and conjuncts, as we don't have enough lexical-semantic information represented to choose the right dependencies in these cases.[4] Secondly, we use a mechanism for hand-coding reading preferences among rules and lexical entries.

---

[4] Consider, for example, genitive modification: The semantic relationship between modifier and modifiee is dependent on their semantic properties: *toukyou no kaigi* - 'the meeting in Tokyo', *watashi no hon* - 'my book'. More lexical-semantic information is needed to choose the correct parse in more complex structures, such as in *watashi no toukyou no imooto* – 'My sister in Tokyo'.

Restrictions like *head-complement preferred to head-adjunct* are quite obvious. Others require domain-specific mechanisms that shall be subject of further work. Stochastic disambiguation methods being developed for the ERG by the Redwoods project at Stanford University (Oepen et al. 2002) should be applicable to this grammar as well.

## 5  Evaluation

The grammar currently covers 93.4% of constructed examples for the banking domain (747 sentences) and 78.2% of realistic email correspondence data (316 sentences), concerning requests for documents. During three months of work, the coverage in the banking domain increased 48.49%. The coverage of the document request data increased 51.43% in the following two weeks.

| Phenomenon | total items # | positive items # | word string % | lexical items Ø | parser analyses Ø | total results # | overall coverage % |
|---|---|---|---|---|---|---|---|
| Total | 747 | 747 | 101 | 75.24 | 6.54 | 698 | 93.4 |

Fig.3 Coverage of banking data, generated by [incr tsdb()]

| Phenomenon | total items # | positive items # | word string % | lexical items Ø | parser analyses Ø | total results # | overall coverage % |
|---|---|---|---|---|---|---|---|
| Total | 316 | 316 | 1.00 | 83.90 | 39.91 | 247 | 78.2 |

Fig.4 Coverage of document request data, generated by [incr tsdb()]

We applied the grammar to unseen data in one of the covered domains, namely the FAQ site of a Japanese bank. The coverage was 61%. 91.2% of the parses output were associated with all well-formed MRSs. That means that we could get correct MRSs in 55.61% of all sentences.

## Conclusion

We described a broad coverage Japanese grammar, based on HPSG theory. It encodes syntactic, semantic, and pragmatic information. The grammar system is connected to a morphological analysis system and uses default entries for words unknown to the HPSG lexicon.

Some basic constructions of the Japanese grammar were described. As the grammar is aimed at working in applications with real-world data, performance and robustness issues are important.

The grammar is being developed in a multilingual context, where much value is placed on parallel and consistent semantic representations. The development of this grammar constitutes an important test of the cross-linguistic validity of the MRS formalism.

The evaluation shows that the grammar is at a stage where domain adaptation is possible in a reasonable amount of time. Thus, it is a powerful resource for linguistic applications for Japanese.

In future work, this grammar could be further adapted to another domain, such as the EDR newspaper corpus (including a headline grammar). As each new domain is approached, we anticipate that the adaptation will become easier as resources from earlier domains are reused. Initial evaluation of the grammar on new domains and the growth curve of grammar coverage should bear this out.